\pdfoutput=1
\documentclass[sigconf]{acmart}

\AtBeginDocument{%
  \providecommand\BibTeX{{%
    \normalfont B\kern-0.5em{\scshape i\kern-0.25em b}\kern-0.8em\TeX}}}

\usepackage{graphicx}
\usepackage{booktabs}
\usepackage{multirow}
\usepackage{enumitem}

\usepackage{amssymb}
\usepackage{amsmath}
\usepackage{bbding}
\usepackage{verbatimbox}
\usepackage{adjustbox}

\usepackage{colortbl} 
\usepackage{xcolor}
\usepackage{pifont}
\usepackage{multirow}

\usepackage{algorithm}

\renewcommand\footnotetextcopyrightpermission[1]{}

\begin{document}

\title{SEVA: Leveraging Single-Step Ensemble of Vicinal Augmentations for Test-Time Adaptation}


\author{Zixuan Hu}
\affiliation{%
  \institution{School of Computer Science, Peking University}
  \city{Beijing}
  \country{China}}
\email{hzxuan@pku.edu.cn}

\author{Yichun Hu}
\affiliation{%
  \institution{School of Computer Science, Peking University}
  \city{Beijing}
  \country{China}}
\email{hycc@pku.edu.cn}

\author{Ling-Yu Duan*}
\affiliation{%
  \institution{School of Computer Science, Peking University}
  \city{Beijing}
  \country{China}}
\email{lingyu@pku.edu.cn}

\begin{abstract}
Test-Time adaptation (TTA) aims to enhance model robustness against distribution shifts through rapid model adaptation during inference. While existing TTA methods often rely on entropy-based unsupervised training and achieve promising results, the common practice of a single round of entropy training is typically unable to adequately utilize reliable samples, hindering adaptation efficiency. In this paper, we discover augmentation strategies can effectively unleash the potential of reliable samples, but the rapidly growing computational cost impedes their real-time application. To address this limitation, we propose a novel TTA approach named Single-step Ensemble of Vicinal Augmentations (SEVA), which can take advantage of data augmentations without increasing the computational burden. Specifically, instead of explicitly utilizing the augmentation strategy to generate new data, SEVA develops a theoretical framework to explore the impacts of multiple augmentations on model adaptation and proposes to optimize an upper bound of the entropy loss to integrate the effects of multiple rounds of augmentation training into a single step. Furthermore, we discover and verify that using the upper bound as the loss is more conducive to the selection mechanism, as it can effectively filter out harmful samples that confuse the model. Combining these two key advantages, the proposed efficient loss and a complementary selection strategy can simultaneously boost the potential of reliable samples and meet the stringent time requirements of TTA. The comprehensive experiments on various network architectures across challenging testing scenarios demonstrate impressive performances and the broad adaptability of SEVA. The code will be publicly available. 
\end{abstract}






\maketitle

\section{Introduction}
Although deep neural network methods have achieved impressive success in many fields \cite{resnet, Non_local}, they heavily rely on the assumption that training and testing data are drawn from the same distribution, a premise that often falters in real-world applications \cite{ben2010theory,vapnik1991principles}. For instance, sensor deterioration can introduce various forms of noise, and natural variations cause fluctuations in weather conditions, among other factors that can markedly alter images \cite{hendrycks2021many,koh2021wilds}. In such deployment scenarios, testing data inevitably encounters distribution shifts, leading to severe performance degradation even with minor shifts \cite{saito2018maximum,hu2024lead,su2020adapting}. This widespread challenge has underscored the crucial need for model adaptation.

Recently, Test-Time adaptation (TTA) methods have emerged and garnered significant attention to address distribution shifts \cite{eata,tent}. Compared to classical domain adaptation methods \cite{ganin2015unsupervised, wang2018deep}, TTA scenarios are more practical: (1) They involve limited access to the entire testing data. (2) Adaptation only occurs at test time and relies on limited backward passes (typically only one pass). These requirements that models should quickly adapt to target distribution with limited information \cite{sar}, pose the key challenge of pursuing high adaptation accuracy and efficiency.

Aiming to improve the accuracy of adaptation using unlabeled data, various sample selection methods have been proposed \cite{eata,sar,rdumb}. Based on indicators like prediction confidence, these methods filter samples that may lead to harmful adaptation directions and only train on reliable samples using entropy loss. While these approaches can capture beneficial direction, they overlook another crucial factor for TTA: how to fully exploit the potential of reliable samples for further improved adaptation performance.

\begin{figure}[t]
\setlength{\abovecaptionskip}{-0.1cm} 
\setlength{\belowcaptionskip}{-0.1cm}
\begin{center}
\includegraphics[width=1.0\linewidth]{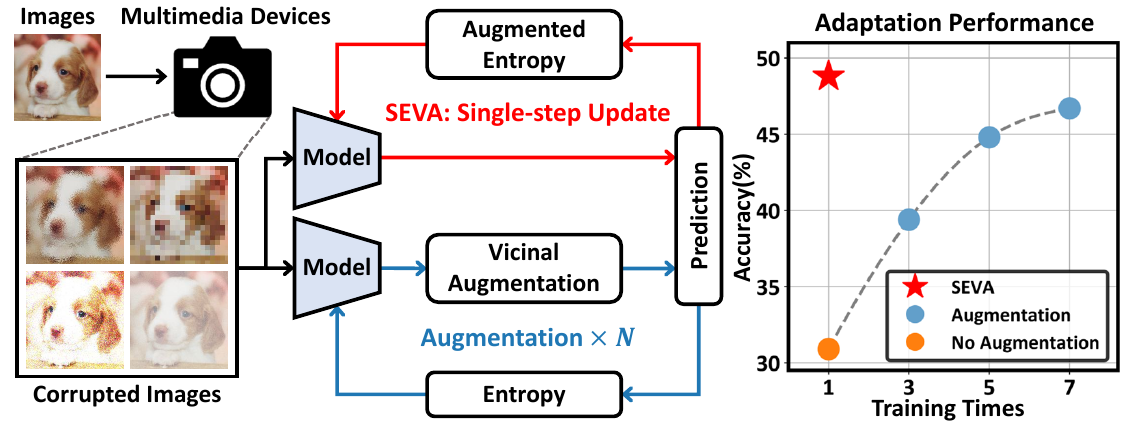}
\end{center}
\caption{\small Multimedia devices often capture corrupted samples, causing distribution shifts that impair model capability. To address this, many TTA methods utilize entropy self-training to adapt the model to these shifts and can use augmented data for multiple training to enhance adaptation, attaining improvements but with significantly increased time overhead. Comparatively, our proposed SEVA achieves the benefits of multi-step augmentation in a single step, yielding impressive results without additional training iterations.}
\label{fig:vit_aug}
\end{figure}

From our observation in Fig. \ref{fig:vit_aug},  repeatedly inputting selected samples into training with data augmentation \cite{noisy_inject} can effectively attain a stable improvement compared to no augmentation, indicating the potential to further enhance adaptation in an augmentation manner. However, despite the benefits for performance, the computational cost of multiple backward updates cannot meet the real-time requirements of TTA testing. This dilemma stems from the need to simultaneously exploit the potential of selected samples and maintain a short adaptation time, making it difficult to enhance adaptation efficiency.

To resolve this issue, we propose a new TTA method that leverages the Single-step Ensemble of Vicinal Augmentations (SEVA) to achieve the effect of multiple rounds of augmentation with only one backward pass, significantly improving adaptation efficiency. Specifically, we introduce the widely adopted vicinal augmentation strategy \cite{noisy_inject,vrm}, which perturbs the original data within the feature space, and analyze its statistical impact on the entropy loss by theoretically deriving the upper bound of the loss in expectation. Rather than directly using the augmentation procedure, our method optimizes the upper bound to implicitly supervise the training on augmented samples. This approach achieves the effect of integrating multiple rounds of impactful augmentation operations into one pass. Therefore, this upper bound can effectively serve as a novel loss named \textit{Augmented Entropy} to accelerate adaptation. Furthermore, building upon Augmented Entropy, we design a selection mechanism that can filter out harmful samples that confuse model prediction, ensuring a reliable adaptation direction. Due to the above designs, our method can significantly enhance adaptation efficiency and achieve stable performance improvements.

To summarize, main contributions of this paper are as follows:
\begin{enumerate}[leftmargin=*]
    \item To the best of our knowledge, we are the first to effectively compress multiple rounds of training into a single round for the TTA scenario. We derive an efficient loss named \textit{Augmented Entropy} that allows one training step to attain a similar effect to numerous rounds of vicinal augmentation training, significantly enhancing the effectiveness of single-step adaptation. 
    \item We analyze and empirically verify advantages of our loss over entropy for selection, as it can effectively filter harmful samples that confuse the model, enabling more reliable selection.
    \item We validate the effectiveness and generalizability of our method through experiments on both ResNet and ViT, achieving SOTA results with an average gain of \textbf{+5.1}, \textbf{+1.5}, \textbf{+3.5} in accuracy across three wild test scenarios of label shifts, mixed domain, and limited batch size, respectively.
\end{enumerate} 
\section{Related Work}
\subsection{Test-Time Adaptation}
Test-Time adaptation aims to enhance the performance on out-of-distribution samples during inference. Depending on whether alter training process, TTA methods can be mainly divided into two groups: 1) Test-Time Training (TTT) \cite{ttt,ttt++,mt3} jointly optimizes the model on training data with both supervised and self-supervised losses, and then conducts self-supervised training at test time. 2) Fully Test-Time Adaptation (Fully TTA) \cite{ctta,tent,mecta} refrains from altering the training process and focuses solely on adapting the model during testing. In this paper, we focus on Fully TTA, as it is more generally applicable than TTT, allowing adaptation of arbitrary pre-trained models without access to training data. 

Due to the wide range of applications for TTA, a variety of methods have been developed \cite{rdumb,eata,sar,bna3,ctta,ecotta}. For instance, some methods adjust the affine coefficients of Batch Normalization layers using testing samples to adapt to distribution shifts \cite{bna1,bna2,bna3}. Others focus on maximizing prediction consistency among different augmented copies of a given sample \cite{memo,chen2022contrastive, s2021test}. DDA \cite{DDA} proposes using a diffusion model for input adaptation to ensure consistency between training and testing distributions. Tent \cite{tent} suggests minimizing the entropy of test samples to increase the confidence of model prediction and reduce error rates. Subsequently, numerous works \cite{zeng2023exploring,yu2023noise} follow this practice of entropy-based training. Building upon Tent, RDumb \cite{rdumb} introduces a strategy in which the model resets its pre-trained weights at regular intervals to reduce the risk of collapse. EATA \cite{eata} and SAR \cite{sar} propose selection strategies to filter out harmful samples and train on selected reliable samples using entropy, thereby enhancing the accuracy of the adaptation process. While this entropy-based training has shown some improvements, we find that it cannot fully utilize the potential of reliable samples, leading to inadequate adaptation efficiency. Instead, our SEVA proposes an Augmented Entropy to unleash the potential, enhancing the efficiency of single-step adaptation.

\subsection{Data Augmentation in TTA}
Data augmentation is an effective technique widely adopted in many fields to artificially expand the size of a dataset by creating modified versions of samples \cite{shorten2019survey,mikolajczyk2018data}. This technique has been evaluated across a variety of tasks, consistently showing its ability to reduce the risk of overfitting and improve model generalization \cite{tran2021data,zoph2020learning,olsson2021classmix}. It was originally applied to the image space by designing various handcrafted transformations like image erasing \cite{zhong2020random,devries2017improved}, image mix \cite{mixup,inoue2018data}, etc. Subsequently, this approach extends to the feature space, allowing for more flexible operations that can be inserted into any position of the network, such as Manifold Mix \cite{verma2019manifold}, Vicinal augmentation \cite{vrm,noisy_inject}, etc.

Naturally, data augmentation is also extensively utilized in TTA methods. The consistency-preserved method \cite{s2021test} proposes minimizing the differences between model predictions for different augmented copies of the same sample, improving the model's robustness to perturbations. MEMO \cite{memo} computes the average of uniformly sampled augmentations to obtain more confident predictions. Subsequently, many works adopt this approach, employing multiple augmentations to average the predictions and devising various variants \cite{wang2024continual,yuan2023robust,ctta}, with the goal of enhancing prediction robustness. For example, CoTTA \cite{ctta} utilizes the average of predictions from 32 augmentations based on a confidence indicator. In contrast to these augmentation-based methods, our approach does not directly generate new samples through augmentation. Instead, we analyze the impact of vicinal augmentations on the loss through theoretical derivation to obtain a novel loss function. By optimizing this loss function, our optimization process can achieve the effect of several rounds of augmentation training in a single step.
\section{Methodology}
\begin{figure*}[t] 
\setlength{\abovecaptionskip}{-0.05cm}
\setlength{\belowcaptionskip}{-0.25cm}
\begin{center}
\includegraphics[width=1.0\linewidth]{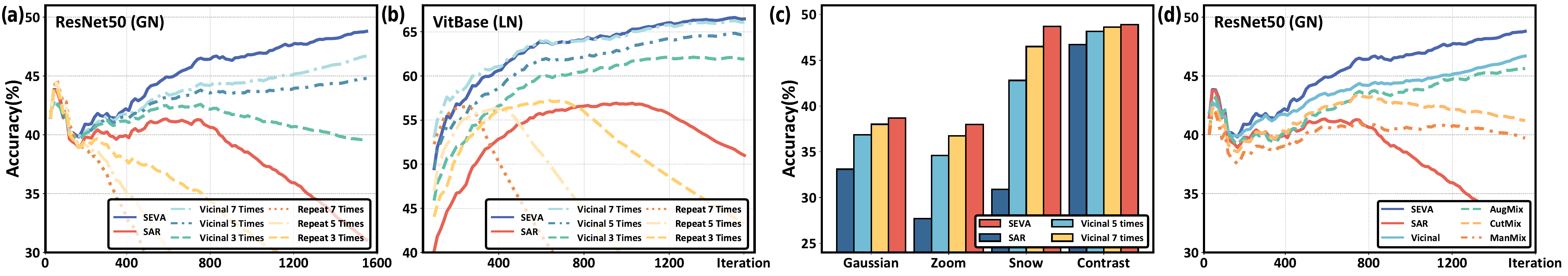}
\end{center}
\caption{\small Performance comparison is conducted on the snow type of ImageNet-C under different training strategies. (a) \& (b) record accuracy during online adaptation and compare SAR \cite{sar}, vicinal augmentation, repeat training, and our SEVA on both ResNet and ViT. "Vicinal/Repeat n Times" denotes utilizing vicinal augmentation/repeat training n times based on SAR. (c) records the final accuracy and compares SEVA with the direct augmentation strategy on four representative corruption types. (d) shows the performance using different augmentation methods.}
\label{fig:compare}
\end{figure*}
\subsection{Problem Setup}
In Test-Time adaptation (TTA), we have a model $F_{\theta}$ that has been pre-trained on the source domain $\mathcal{D}_{\mathcal{S}} = \{X_{train}^{i}, Y_{train}^i\}_{i=1}^{|\mathcal{S}|}$ and need to evaluate it on the target domain $\mathcal{D}_{\mathcal{T}}=\{X_{test}^i, Y_{test}^i\}_{i=1}^{|\mathcal{T}|}$, where $\theta$ denotes the model parameters, $X, Y$ denote samples and labels, and $|\mathcal{S}|, |\mathcal{T}|$ denote the number of training data and testing data, respectively. Due to the distribution shift $P(X_{train}, Y_{train})\neq P(X_{test}, Y_{test})$, Test-Time adaptation requires the model $F_{\theta}$ to adapt to the target domain without ground-truth labels $Y_{test}$. Therefore, most existing TTA methods \cite{tent,eata,sar} update $\theta$ through self-training, minimizing the entropy loss $\mathcal{L}_{ent}$ on $X_{test}$:
\begin{equation}
    \mathcal{L}_{ent}(x)=-p_{\theta}(x) \cdot \log p_{\theta}(x)=-\sum_{i=1}^C p_{\theta}(x)_i \log p_{\theta}(x)_i,
\label{eq:entropy}
\end{equation}
where $C$ denotes the number of classes and $p_{\theta}(x)=F_{\theta}(x)=\left(p_{\theta}(x)_1, \ldots, p_{\theta}(x)_C\right) \in \mathbb{R}^C$ denotes the predicted probabilities. 

\subsection{Exploring the Potential of Reliable Samples}

Reliable samples can provide trustworthy guidance for model adaptation, alleviating the model collapse issue in TTA \cite{eata,sar}. Despite these benefits, the potential of these samples to further promote adaptation remains under-explored due to inadequate training processes in TTA. For instance, while data augmentation has proved a powerful technique that can unleash the potential of training data in other transfer learning tasks through enriching individual samples into diverse sets ~\cite{volpi2018generalizing,huang2018auggan}, its capabilities and mechanisms on Test-Time adaption are yet to be comprehensively studied.

In this paper, we empirically test the impact of data augmentation on the TTA task under different models and various domain shifts. We select two widely adopted image augmentations (AugMix \cite{augmix}, CutMix \cite{cutmix}) and two feature augmentations (Manifold Mix \cite{verma2019manifold}, Vicinal augmentation \cite{noisy_inject,vrm}) for evaluation. During the online adaptation process, we train models on reliable samples selected by SAR \cite{sar} using different augmentation strategies and record the accuracy changes of ResNet50-GN \cite{resnet} and ViT-LN \cite{ViT} in Fig. \ref{fig:compare}.

Comparing the results in Fig. \ref{fig:compare}, we make two key observations. First, from Fig. \ref{fig:compare}(a) \& (b), we observe that repeatedly training with original samples is unstable and leads to significant degradation. With an increasing number of repetitions, the risk of overfitting grows, accelerating performance collapse and ultimately resulting in adaptation failure. In contrast, training with data augmentation can significantly reduce the risk of overfitting and achieve stable improvements. Second, as shown in Fig. \ref{fig:compare}(d), we notice the superior performance of vicinal augmentation. This augmentation, as demonstrated in many studies, introduces richer semantic changes compared to other augmentations, as it perturbs the feature space in all directions \cite{lim2022noisy}. In Fig. \ref{fig:compare}(c), we test vicinal augmentation in various distribution shifts and observe a consistent improvement compared to no augmentation, further highlighting the potential of reliable samples to enhance adaptation.

The observations encourage us to employ simple and effective augmentation techniques in TTA to fully utilize selected samples. However, repeatedly training samples with augmentation introduces a several-fold increase in time overhead, which cannot meet the real-time requirement of TTA. Therefore, we hope to propose a method that can achieve a similar effect to augmentation in promoting adaptation while compressing the computational cost of multiple training rounds.

\subsection{Single Training Integrates Multiple Vicinal Augmentations}
\label{sec:method3.3}

\begin{figure*}[t]
\setlength{\abovecaptionskip}{-0.03cm}
\setlength{\belowcaptionskip}{-0.27cm}
\begin{center}
\includegraphics[width=1.0\textwidth]{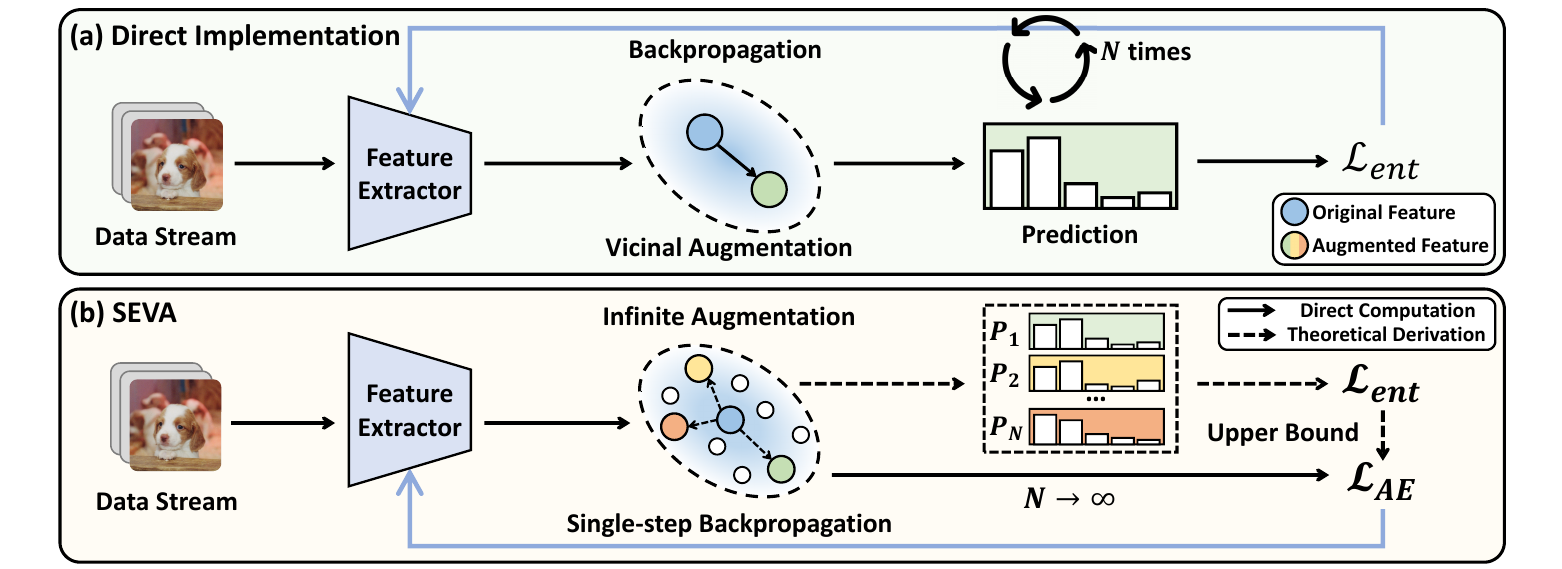}
\end{center}
\caption{\small Comparison of different approaches for using vicinal augmentation in the TTA scenario: (a) Direct Augmentation: It involves repeatedly sampling augmented features into training from a Gaussian distribution (as the vicinal area). As the number of repetitions $N$ increases, augmented sets become richer, but the computational costs also increase linearly and cannot meet the time requirement. (b) Efficient Approach in SEVA: Instead of directly augmenting samples, we consider the impact of vicinal augmentation on the entropy loss as the number of augmentations approaches infinity. Through theoretical derivation, we obtain a closed-form upper bound to serve as a novel loss $\mathcal{L}_{AE}$ and leverage a single training step of $\mathcal{L}_{AE}$ to achieve a similar effect to multiple rounds in (a), significantly enhancing the adaptation efficiency.}
\label{fig:pipeline}
\end{figure*}

To address the aforementioned challenge, our study delves deeper into analyzing the impact of data augmentation on entropy loss. Since vicinal augmentation showcases superior performance in Fig. \ref{fig:compare}(d), and has a concise formula that facilitates our theoretical framework, we choose to analyze its role as the augmentation strategy on entropy loss. Through theoretical derivation, we obtain the probability distribution of the model output and a closed-form upper bound of the loss under the effects of random augmentations. Based on these results, we propose a new TTA method called "Single-step Ensemble of Vicinal Augmentations (SEVA)", which consists of two important components: an efficient loss called \textit{Augmented Entropy} that achieves similar effects to multiple rounds of augmentation in a single training step and a reliable sample selection mechanism to filter out harmful samples that confuse adaptation.

In this subsection, we first obtain the probability distribution of model outputs influenced by vicinal augmentation, as shown in Eq. \ref{eq:prob_aug}. Furthermore, by considering an infinite number of vicinal augmentations, we can derive a closed-form upper bound for the entropy loss, as given in Eq. \ref{eq:loss}, which can be treated as a novel loss and enables efficient adaptation. Finally, we propose to set the boundary of the new loss for sample filtering in Eq. \ref{eq:minimization} and compare its advantages over selection based on vanilla entropy.

\subsubsection{Robust Prediction under Augmentation.} 
In our SEVA, we consider the effect of vicinal augmentation on the feature extracted by the backbone, which directly affects the probability distribution of predictions. Considering a sample $x$ and its corresponding feature $z\in R^{d}$, the probability outputted by the classifier is given by:
\begin{equation}
    p_{\theta}(z)_i = \left(\text{softmax} \left(Az+b\right)\right)_i=\frac{e^{a_i\cdot z+b_i}}{\sum_{j=1}^{C}e^{a_j\cdot z +b_j}}, i=1, 2,\ldots, C,
\label{eq:prob}
\end{equation}
where the subscript $\left(\cdot\right)_i$ represents the $i$-th dimension, corresponding to the $i$-th class. $C$ denotes the number of classes. $A\in \mathbb{R}^{C\times d}$ and $b\in \mathbb{R}^{C}$ are the linear and bias coefficients of the affine layer in the classifier, with $a_i$ and $b_i$ being their $i$-th dimensions, respectively.

During vicinal augmentation, we treat the vicinal range of a sample $z$ in the feature space as a Gaussian distribution $\mathcal{N}(z, \Sigma)$ and generate new data $\widetilde{z}$ by Monte-Carlo sampling from this distribution, where $\Sigma$ is a predefined covariance matrix that determines the range of the vicinal area. Previous studies \cite{naveed2024survey, li2023modelinguncertainfeaturerepresentation} have demonstrated that perturbations introduced in this manner can often lead to rich semantic variations, which can affect the predicted probabilities of model outputs. As these probability changes reflect the model's uncertainty when encountering corruption, we calculate the expectation of outputs to incorporate the effects of perturbations in all directions, achieving more robust predictions:
\begin{equation}
\begin{aligned}
\overline{p_{\theta}(z)}_i :=\frac{\mathbb{E}_{\tilde{z} \sim \mathcal{N}(z, \Sigma)} e^{a_i \cdot \tilde{z}+b_i}}{\mathbb{E}_{\tilde{z} \sim \mathcal{N}(z, \Sigma)} \sum_{j=1}^C e^{a_j \cdot \tilde{z}+b_j}}  =\frac{e^{a_i \cdot z+b_i+\frac{1}{2} a_i \Sigma a_i^{\top}}}{\sum_{j=1}^C e^{a_j \cdot z+b_j+\frac{1}{2} a_j \Sigma a_j^{\top}}},
\end{aligned}
\label{eq:prob_aug}
\end{equation}
where $\mathbb{E}$ denotes the operation of expectation. Notably, to ensure that the obtained probabilities satisfy the basic requirement of summing to 1, we choose to calculate the expectation before normalization, instead of after. The second equality in Eq. \ref{eq:prob_aug} is derived by leveraging the moment property of the Gaussian distribution $\mathbb{E}_{X\sim\mathcal{N}(\mu, \sigma^2)}e^{X}=e^{\mu+1/2 \sigma^2}$ and $a_i\cdot \tilde{z}+b_i\sim \mathcal{N}\left(a_i
\cdot z+b_i, a_i \Sigma a_i^{\top}\right)$. 
Through theoretical derivation, we integrate the influence of augmentation in various directions to obtain $\overline{p_{\theta}(z)}$ as our desired robust prediction, enabling further exploration of the impact of vicinal augmentations on model optimization.

\subsubsection{Single Step to Integrate Augmentations.}
After studying the impact of vicinal augmentation on the output probabilities, we further analyze its effect on the entropy loss. We first consider the case of a finite number of augmentations, where a single feature $z$ is augmented $N$ times, forming augmented features $\tilde{z}_1, \tilde{z}_2, \ldots, \tilde{z}_N$, which are drawn from the same distribution $\mathcal{N}(z, \Sigma)$ independently, as shown in Fig. \ref{fig:pipeline}(a). Then the model is trained by minimizing the loss shown below:
\begin{equation}
    \begin{aligned}
        \mathcal{L}_{N} &= - \frac{1}{N} \sum_{i=1}^{N}p_{\theta}(\tilde{z}_i) \log p_{\theta}(\tilde{z}_i) =-\frac{1}{N}\sum_{i=1}^{N}\sum_{j=1}^{C}p_{\theta}(\tilde{z}_i)_j \log p_{\theta}(\tilde{z}_i)_j \\
        &=-\frac{1}{N}\sum_{i=1}^{N}\sum_{j=1}^{C}\frac{e^{a_j\cdot \tilde{z}_i+b_j}}{\sum_{k=1}^{C}e^{a_k\cdot \tilde{z}_i +b_k}}\cdot \log \frac{e^{a_j\cdot \tilde{z}_i+b_j}}{\sum_{k=1}^{C}e^{a_k\cdot \tilde{z}_i +b_k}}.
    \end{aligned}
\label{eq:naive}
\end{equation}

This direct approach introduces a linear increase in computational cost with the number of augmentations. When $N$ is large, the feature set enriched by augmentation can provide diverse variations to enhance adaptation, but it fails to meet the time requirement of TTA. To address this, we consider the case where $N$ grows to infinity in Fig. \ref{fig:pipeline}(b) and derive a closed-form upper bound as a new training loss, achieving the goal of highly efficient implementation:
\begin{equation}
\begin{aligned}
\mathcal{L}_{\infty} =&-\lim _{N \rightarrow+\infty} \frac{1}{N} \sum_{i=1}^N \sum_{j=1}^C p_\theta\left(\tilde{z}_i\right)_j \log p_\theta\left(\tilde{z}_i\right)_j \\
 \leqslant&-\lim _{N \rightarrow+\infty} \frac{1}{N} \sum_{i=1}^N \sum_{j=1}^C (\frac{1}{N}\sum_{k=1}^N p_{\theta}(\tilde{z}_k)_j) \log P_\theta\left(\tilde{z}_i\right)_j \\
=&-\sum_{j=1}^C \frac{e^{a_j \cdot z+b_j+\frac{1}{2} a_j \sum a_j{ }^{\top}}}{\sum_{k=1}^C e^{a_k \cdot z+b_k+\frac{1}{2} a_k \sum a_k^{\top}}} \mathbb{E}_{\tilde{z} \sim N(z, \Sigma)} \log p_\theta(\tilde{z})_j \\
\leqslant&\sum_{j=1}^C \frac{e^{a_j \cdot z+b_j+\frac{1}{2} a_j \sum a_j{ }^{\top}}}{\sum_{k=1}^C e^{a_k \cdot z+b_k+\frac{1}{2} a_k \sum a_k^{\top}}} \log \sum_{i=1}^C e^{\left(a_i-a_j\right)\cdot z} \\
& \cdot e^{\left(b_i - b_j\right)+\frac{1}{2}(a_j-a_i)\Sigma(a_j-a_i)^{\top}} \triangleq \mathcal{L}_{AE} \, ,
\end{aligned}
\label{eq:loss}
\end{equation}
where two inequalities are from the Theorem. 1 \& 2 in the appendix, and the equality is derived from the definition of expectation and the robust prediction we obtained in Eq. \ref{eq:prob_aug}. We refer to the final result $\mathcal{L}_{AE}$ as \textit{Augmented Entropy}. 

Through in-depth exploration, we obtain an upper bound of the entropy loss with an infinite number of vicinal augmentations in expectation. Compared to the naive implementation in Eq. \ref{eq:naive}, using the upper bound in Eq. \ref{eq:loss} to optimize the model has two advantages: efficiency and comprehensiveness. On the one hand, it can be directly computed from the feature $z$ without sampling, eliminating the need to increase the number of training iterations for augmented features. On the other hand, it implicitly samples from all directions within the vicinal range, providing more comprehensive guidance for adaptation. Therefore, we have effectively proposed a novel loss that can integrate multiple vicinal augmentations to replace the original entropy loss, offering a more efficient adaptation.

\subsubsection{Reliable Augmented Entropy Minimization.}
\label{sec:selection}
\noindent Previous studies have shown that samples with high entropy often lead to large and noisy gradients during adaptation, thereby using an entropy boundary to filter out such samples \cite{sar,rdumb}. We discover that the proposed Augmented Entropy can more sensitively filter harmful samples that confuse the model prediction compared to vanilla entropy. Leveraging this advantage, we employ the boundary condition of Augmented Entropy instead of entropy as the complementary selection strategy to identify reliable samples for training with this new loss. Formally, the optimization of our reliable Augmented Entropy minimization is defined as:

\begin{equation}
    \min _{\theta} R(x) \mathcal{L}_{AE}(x), \text { where } R(x) \triangleq \mathbb{I}_{\left\{\mathcal{L}_{AE}(x)<\mathcal{L}_0\right\}}(x),
\label{eq:minimization}
\end{equation}
where $\mathbb{I}$ is an indicator function and $\mathcal{L}_0$ is a pre-defined boundary.

During adaptation, the model often encounters confusion caused by corrupted samples, leading to difficulty in distinguishing between the ground-truth i-th class and the erroneous j-th class. In such cases, training based on entropy will adopt highly similar optimization for the probabilities of these two classes, intensifying the confusion between them. When the two classes are similar, the entropy-based training causes a slight effect. However, for the vastly different classes, it can blur classification boundaries and induce severe noises for model adaptation. Therefore, during sample selection, we aim to exclude samples for which the predicted probabilities of two classes are similar but these two classes have significant differences. To analyze the advantage of our loss mentioned above, we transform two losses into a similar form and it becomes evident that $\mathcal{L}_{AE}$ has a crucial modification: a weight term $e^{\frac{1}{2}\left(a_j-a_i\right) \sum\left(a_j-a_i\right)^{\top}}$ is added.

\begin{equation}
\setlength\abovedisplayskip{-6pt}
\setlength\belowdisplayskip{1pt}
\begin{aligned}
& \mathcal{L}_{ent}=\sum_{j=1}^C p_\theta(z)_j \cdot\log \sum_{i=1}^C \frac{p_\theta(z)_i}{p_\theta(z)_j}, \\
& \mathcal{L}_{AE}=\sum_{j=1}^C \overline{ p_\theta(z)}_j \cdot\log \sum_{i=1}^C \frac{p_\theta(z)_i}{p_{\theta}(z)_j}\cdot e^{\frac{1}{2}\left(a_j-a_i\right) \sum\left(a_j-a_i\right)^{\top}}.
\end{aligned}
\label{eq:two_loss}
\end{equation}

Observed from Eq. \ref{eq:two_loss}, the vanilla entropy lacks the ability to distinguish the distance between different classes and cannot effectively exclude such harmful cases. In contrast, our Augmented Entropy incorporates the weight $e^{\frac{1}{2}\left(a_j-a_i\right) \sum\left(a_j-a_i\right)^{\top}}$ that measures the distance between different classes. In fact, $a_{1}, \ldots, a_{C}$ as the affine parameters of the classifier represent the prototypes of each class, and the distance between them can effectively reflect the similarity between classes. For samples with similar probabilities $p_\theta(z)_i\approx p_\theta(z)_j$ and large distance $||a_j-a_i||_2$, the weight can significantly increase the overall Augmented Entropy, making them more likely to be excluded. Therefore, our Augmented Entropy can effectively filter out harmful samples which can confuse the model, providing more reliable samples for training.

\subsubsection{Summary of Our Method.}
We explore the substantial potential of reliable samples for enhancing adaptation in an augmentation manner and find that vicinal augmentation can effectively facilitate adaptation. Through an in-depth exploration of the impact of vicinal augmentation on the model prediction and entropy loss, we derive the efficient Augmented Entropy to integrate the effects of multiple vicinal augmentations in a single step, fully leveraging the potential of samples and improving adaptation efficiency. Furthermore, we propose a selection mechanism coupled with Augmented Entropy to select more reliable samples for training. Based on the novel training loss function and complementary selection mechanism, our SEVA can more stably select reliable training samples and significantly improve the efficiency of adaptation.
\vspace{-0.1cm}
\section{EXPERIMENTS}
\vspace{-0.03cm}
To demonstrate the effectiveness and robustness of our approach, we evaluate our SEVA on wild TTA benchmark proposed in \cite{sar}, including three real-world test scenarios, \textit{i.e.},  imbalanced label shifts, mixed testing domain, and limited batch size. In addition, we provide a runtime comparison to verify the efficiency of SEVA.
\vspace{-0.15cm}
\subsection{Experimental Details}
\subsubsection{Benchmark and Test Scenarios.}
We conduct our experiments on ImageNet-C \cite{imagenet-c}, a large-scale dataset for testing the robustness of models against common domain shifts in the real world. Compared to the original ImageNet dataset \cite{imagenet}, ImageNet-C contains images with various visual perturbations and distortions added artificially. It includes 15 types of corruption across 4 main categories (noise, blur, weather, digital), and each type has 5 severity levels, ranging from level 1 to level 5, indicating the increasing severity of corruption. We evaluate and compare different methods in three challenging wild scenarios: 1) The label distribution of the testing data is online shifted and imbalanced at each time step. 2) The input distribution of the testing data is a mixture of all types, and the model will encounter continuously changing corruptions. 3) Limited batch size that model adaptation can only capture extremely limited information at each iteration.

\subsubsection{Models and Implementation Details.}
Regarding the choice of models, as previous work \cite{sar} has shown that batch-agnostic normalization layers (i.e., GN \cite{groupnormalization} and LN \cite{layernorm}) are more beneficial than BN \cite{batchnorm} for stable test-time adaptation, we conduct experiments on ResNet50-GN \cite{resnet} and ViTBase-LN \cite{ViT} obtained from \texttt{timm} \cite{rw2019timm}. For the optimizer, we follow SAR \cite{sar} to use SGD, batch size of 64 (except for experiments with limited batch size=1), with a momentum of 0.9, and a learning rate of 0.00025/0.001 for ResNet/ViT. For SEVA, the boundary $\mathcal{L}_0$ in Eq. \ref{eq:minimization} is set to $1.0\times \text{ln}1000$ and $\Sigma$ in Eq. \ref{eq:loss} is set to $\lambda\cdot\Sigma_{\mathcal{T}}$ where $\Sigma_{\mathcal{T}}$ is set to a fixed value which is the variance of a small subset (128 samples) of features from the testing data and $\lambda$ is $1.5$ by default. For trainable parameters, according to common practices \cite{tent,sar}, we adapt the affine parameters of normalization layers. 

\begin{table*}[t]
\setlength{\abovecaptionskip}{-0.03cm} 
\caption{Comparisons with state-of-the-art methods on ImageNet-C (severity level 5) under \textbf{Imbalanced Label Shift} regarding Accuracy (\%). We report mean\&std over 3 independent runs. The best results are in bold and the second-best are in underline.}
\centering
\label{tab:label shift}
\resizebox{\textwidth}{!}{
\fontsize{17}{20}\selectfont
\begin{tabular}{c|ccc|cccc|cccc|cccc|c}
\toprule[1pt]
\multirow{2}{*}{\centering\textbf{Model+Method}}      & \multicolumn{3}{c|}{Noise}              & \multicolumn{4}{c|}{Blur}                          & \multicolumn{4}{c|}{Weather}                       & \multicolumn{4}{c|}{Digital}                       &     \multirow{2}{*}{\centering\textbf{Average}}       \\
    & Gauss.       & Shot       & Impul.     & Defoc.     & Glass      & Motion     & Zoom       & Snow       & Frost      & Fog        & Brit.      & Contr.     & Elastic    & Pixel      & JPEG       &         \\
\toprule[1pt]
\multicolumn{1}{c|}{ResNet50(GN)}                                                     & 17.9         & 19.9       & 17.9       & \underline{19.7}       & 11.3       & 21.3       & 24.9       & 40.4       & \underline{47.4}       & 33.6       & 69.2       & 36.3       & 18.7       & 28.4       & 52.2       & 30.6       \\
\multicolumn{1}{l|}{\quad$\bullet$ MEMO}       & 18.4         & 20.6       & 18.4       & 17.1       & 12.7       & 21.8       & 26.9       & \underline{40.7}       & 46.9       & 34.8       & 69.6       & 36.4       & 19.2       & 32.2       & 53.4       & 31.3       \\
\multicolumn{1}{l|}{\quad$\bullet$ DDA}          & \textbf{42.5}         & \textbf{43.4}       & \textbf{42.3}       & 16.5       & 19.4       & 21.9       & 26.1       & 35.8       & 40.2       & 13.7       & 61.3       & 25.2       & \textbf{37.3}       & 46.9       & 54.3       & 35.1       \\
\multicolumn{1}{l|}{\quad$\bullet$ Tent}       & 2.6          & 3.3        & 2.7        & 13.9       & 7.9        & 19.5       & 17.0       & 16.5       & 21.9       & 1.8        & 70.5       & 42.2       & 6.6        & 49.4       & 53.7       & 22.0       \\
\multicolumn{1}{l|}{\quad$\bullet$ EATA}     & 27.0         & 28.3       & 28.1       & 14.9       & 17.1       & 24.4       & 25.3       & 32.2       & 32.0       & 39.8       & 66.7       & 33.6       & \underline{24.5}       & 41.9       & 38.4       & 31.6       \\
\multicolumn{1}{l|}{\quad$\bullet$ EcoTTA}    & 31.7         & 24.5       & 27.4       & 16.7       & 17.3       & 26.0       & 27.2       & 35.4       & 41.4       & 45.8       & 69.6       & 43.2       & 8.2       & 40.2       & 50.8       & 33.7       \\
\multicolumn{1}{l|}{\quad$\bullet$ RDumb}       & 32.3         & 34.8       & 31.3       & 18.9       & \underline{19.8}       & \underline{35.2}       & \underline{29.3}       & 34.9       & 38.0       & 48.5       & 70.5       & 43.3       & 9.0       & 47.0       & 48.7       & 36.1       \\
\multicolumn{1}{l|}{\quad$\bullet$ SAR}         & 33.1         & 36.5       & 35.5       & 19.2       & 19.5       & 33.3       & 27.7       & 23.9       & 45.3       & \underline{50.1}      & \underline{71.9}       & \underline{46.7}       & 7.1        & \underline{52.1}       & \underline{56.3}       & \underline{37.2}       \\
\rowcolor[HTML]{EBF8FF}
\multicolumn{1}{l|}{\quad$\bullet$ SEVA (Ours)} & 
$\underline{38.7}_{\pm 0.2}$ & 
$\underline{40.8}_{\pm 0.2}$ & 
$\underline{39.5}_{\pm 0.3}$ & 
$\textbf{21.7}_{\pm 0.9}$ & 
$\textbf{23.2}_{\pm 0.5}$ & 
$\textbf{37.1}_{\pm 0.2}$ & 
$\textbf{38.0}_{\pm 0.2}$ & 
$\textbf{48.7}_{\pm 3.2}$ & 
$\textbf{48.1}_{\pm 0.2}$ & 
$\textbf{53.5}_{\pm 0.9}$ & 
$\textbf{72.5}_{\pm 0.1}$ & 
$\textbf{48.7}_{\pm 0.3}$ & 
$8.7_{\pm 2.3}$ & 
$\textbf{54.9}_{\pm 0.3}$ & 
$\textbf{57.2}_{\pm 0.3}$ & 
$\textbf{42.1}_{\pm 0.6}$ \\
\toprule[1pt]
ViT(LN)                                                                               & 9.4          & 6.7        & 8.3        & 29.1       & 23.4       & 34.0       & 27.0       & 15.8       & 26.3       & 47.4       & 54.7       & 43.9       & 30.5       & 44.5       & 47.6       & 29.9       \\
\multicolumn{1}{l|}{\quad$\bullet$ MEMO}       & 21.6         & 17.4       & 20.6       & 37.1       & 29.6       & 40.6       & 34.4       & 25.0       & 34.8       & 55.2       & 65.0       & 54.9       & 37.4       & 55.5       & 57.7       & 39.1       \\
\multicolumn{1}{l|}{\quad$\bullet$ DDA}        & 41.3         & 41.3       & 40.6       & 24.6       & 27.4       & 30.7       & 26.9       & 18.2       & 27.7       & 34.8       & 50.0       & 32.3       & 42.2       & 52.5       & 52.7       & 36.2       \\
\multicolumn{1}{l|}{\quad$\bullet$ Tent}       & 32.7         & 1.4        & 34.6       & 54.4       & 52.3       & 58.2       & 52.2       & 7.7        & 12.0       & 69.3       & 76.1       & 66.1       & 56.7       & 69.4       & 66.4       & 47.3       \\
\multicolumn{1}{l|}{\quad$\bullet$ EATA}       & 35.9         & 34.6       & 36.7       & 45.3       & 47.2       & 49.3       & 47.7       & \underline{56.5}       & 55.4       & 62.2       & 72.2       & 21.7       & 56.2       & 64.7       & 63.7       & 50.0       \\
\multicolumn{1}{l|}{\quad$\bullet$ RDumb}       & 42.5         & \underline{48.2}       & 43.1       & 53.1       & 49.9       & 51.8       & 54.7       & 57.6       & \underline{56.7}       & 68.6       & 74.6       & 58.0       & 59.3       & 67.2       & 63.0       & 56.5       \\
\multicolumn{1}{l|}{\quad$\bullet$ SAR}        & \underline{46.5}         & 43.1       & \underline{48.9}       & \underline{55.3}       & \underline{54.3}       & \underline{58.9}       & \underline{54.8}       & 53.6       & 46.2       & \underline{69.7}       & \underline{76.2}       & \underline{66.2}       & \underline{60.9}       & \underline{69.6}       & \underline{66.6}       & \underline{58.1}       \\
\rowcolor[HTML]{EBF8FF}
\multicolumn{1}{l|}{\quad$\bullet$ SEVA (Ours)} & 
$\textbf{52.2}_{\pm 0.5}$ & 
$\textbf{51.9}_{\pm 1.5}$ & 
$\textbf{53.1}_{\pm 0.3}$ & 
$\textbf{57.5}_{\pm 0.3}$ & 
$\textbf{57.9}_{\pm 0.4}$ & 
$\textbf{62.3}_{\pm 0.1}$ & 
$\textbf{59.3}_{\pm 0.2}$ & 
$\textbf{66.3}_{\pm 0.5}$ & 
$\textbf{65.0}_{\pm 0.1}$ & 
$\textbf{72.4}_{\pm 0.3}$ & 
$\textbf{77.4}_{\pm 0.2}$ & 
$\textbf{67.7}_{\pm 0.2}$ & 
$\textbf{66.6}_{\pm 0.3}$ & 
$\textbf{72.3}_{\pm 0.1}$ & 
$\textbf{69.2}_{\pm 0.4}$ & 
$\textbf{63.4}_{\pm 0.4}$ \\
\bottomrule[1pt]
\end{tabular}
}
\end{table*}

\begin{table}[b]
\setlength{\abovecaptionskip}{0cm} 
\setlength{\belowcaptionskip}{-0.5cm}
\begin{center}
\caption{\small Comparisons with state-of-the-art methods on ImageNet-C (severity level 5, 4, \& 3) under Mixture of 15 Corruption Types.}
\label{table:mixed}
\resizebox{\linewidth}{!}{
\begin{tabular}{c|ccc|c}
\toprule[1pt]
  Model+Method   & Level=5    & Level=4    & Level=3    & Average        \\ \hline
ResNet50 (GN) & 30.6       & 42.7       & 54.0       & 42.4       \\
\multicolumn{1}{l|}{\quad$\bullet$ MEMO}         & 31.2       & 43.0       & 54.5       & 42.9       \\
\multicolumn{1}{l|}{\quad$\bullet$ DDA}           & 35.1       & 43.6       & 52.3       & 43.7       \\
\multicolumn{1}{l|}{\quad$\bullet$ TENT}         & 13.4       & 20.6       & 33.1       & 22.4       \\
\multicolumn{1}{l|}{\quad$\bullet$ EATA}        & 38.1       & 47.7       & 56.1       & 47.3       \\
\multicolumn{1}{l|}{\quad$\bullet$ EcoTTA}      & 31.5       & 45.1       & 55.6       & 44.1       \\
\multicolumn{1}{l|}{\quad$\bullet$ RDumb}       & \underline{38.4}       & 47.5       & 55.8       & 47.2       \\
\multicolumn{1}{l|}{\quad$\bullet$ SAR}          & 38.3       & \underline{48.6}       & \underline{57.3}       & \underline{48.1}       \\ 
\rowcolor[HTML]{EBF8FF}
\multicolumn{1}{l|}{\quad$\bullet$ SEVA (Ours)}        & $\textbf{40.0}_{\pm 0.1} $& $\textbf{49.8}_{\pm 0.0} $& $\textbf{58.1}_{\pm 0.1} $& $\textbf{49.3}_{\pm 0.1} $\\ 
\toprule[1pt]
ViT (LN)      & 29.9       & 42.9       & 53.8       & 42.2       \\
\multicolumn{1}{l|}{\quad$\bullet$ MEMO}         & 39.1       & 51.3       & 62.1       & 50.8       \\
\multicolumn{1}{l|}{\quad$\bullet$ DDA}          & 36.1       & 45.1       & 53.2       & 44.7       \\
\multicolumn{1}{l|}{\quad$\bullet$ TENT}         & 16.5       & 64.3       & 70.2       & 50.3       \\
\multicolumn{1}{l|}{\quad$\bullet$ EATA}         & 55.7       & 63.7       & 69.6       & 63.0       \\
\multicolumn{1}{l|}{\quad$\bullet$ RDumb}        & 56.6       & 64.2       & 69.9       & 63.6       \\
\multicolumn{1}{l|}{\quad$\bullet$ SAR}          & \underline{57.1}       & \underline{64.9}       & \underline{70.7}       & \underline{64.2}       \\ 
\rowcolor[HTML]{EBF8FF} \multicolumn{1}{l|}{\quad$\bullet$ SEVA (Ours)}       & $\textbf{59.3}_{\pm 0.0} $ & $\textbf{66.6}_{\pm 0.1} $& $\textbf{71.8}_{\pm 0.1}$ & $\textbf{65.9}_{\pm 0.1} $\\
\toprule[1pt]
\end{tabular}}
\end{center}
\end{table}

\begin{table*}[t]
\setlength{\abovecaptionskip}{-0.03cm} 
\setlength{\belowcaptionskip}{-0.2cm}
\caption{Comparisons with state-of-the-art methods on ImageNet-C (severity level 5) under Limited Batch Size $=$ 1.}
\centering
\label{tab:bs1}
\resizebox{\textwidth}{!}{
{
\fontsize{17}{20}\selectfont
\begin{tabular}{c|ccc|cccc|cccc|cccc|c}
\toprule[1pt]
\multirow{2}{*}{\centering\textbf{Model+Method}}      & \multicolumn{3}{c|}{Noise}              & \multicolumn{4}{c|}{Blur}                          & \multicolumn{4}{c|}{Weather}                       & \multicolumn{4}{c|}{Digital}                       &     \multirow{2}{*}{\centering\textbf{Average}}       \\
    & Gauss.       & Shot       & Impul.     & Defoc.     & Glass      & Motion     & Zoom       & Snow       & Frost      & Fog        & Brit.      & Contr.     & Elastic    & Pixel      & JPEG       &         \\
\toprule[1pt]
\multicolumn{1}{c|}{ResNet50(GN)}                                                      & 18.0       & 19.8         & 17.9         & \textbf{19.8}       & 11.4         & 21.4         & 24.9         & 40.4       & \textbf{47.3}         & 33.6         & 69.3         & 36.3       & 18.6         & 28.4         & 52.3         & 30.6      \\
\multicolumn{1}{l|}{\quad$\bullet$ MEMO}       & 18.5       & 20.5         & 18.4         & 17.1       & 12.6         & 21.8         & 26.9         & 40.4       & \underline{47.0}         & 34.4         & 69.5         & 36.5       & 19.2         & 32.1         & 53.3         & 31.2      \\
\multicolumn{1}{l|}{\quad$\bullet$ DDA}          & \textbf{42.4}       & \textbf{43.3}         & \textbf{42.3}         & 16.6       & \textbf{19.6}         & 21.9         & 26.0         & 35.7       & 40.1         & 13.7         & 61.2         & 25.2       & \textbf{37.5}         & 46.6         & 54.1         & 35.1       \\
\multicolumn{1}{l|}{\quad$\bullet$ Tent}       & 2.5        & 2.9          & 2.5          & 13.5       & 3.6          & 18.6         & 17.6         & 15.3       & 23.0         & 1.4          & 70.4         & 42.2       & 6.2          & \underline{49.2}         & 53.8         & 21.5      \\
\multicolumn{1}{l|}{\quad$\bullet$ EATA}     & 24.8       & 28.3         & 25.7         & 18.1       & 17.3         & 28.5         & 29.3         & 44.5       & 44.3         & \textbf{41.6}         & 70.9         & \underline{44.6}       & \underline{27.0}         & 46.8         & 55.7         & \underline{36.5}     \\
\multicolumn{1}{l|}{\quad$\bullet$ EcoTTA}    & 19.8       & 20.6         & 23.8         & 10.2       & 14.6         & 20.3         & 22.2         & 37.8       & 36.3         & 29.7         & 71.3         & 43.4       & 8.9          & 47.5         & 52.4         & 30.6       \\
\multicolumn{1}{l|}{\quad$\bullet$ RDumb}      & 21.0       & 23.9         & 21.9         & 18.8       & 13.7         & 26.6         & 28.6         & 42.3       & 46.2         & 31.6         & 69.3         & 41.6       & 20.7         & 42.6         & 52.5         & 33.4    \\
\multicolumn{1}{l|}{\quad$\bullet$ SAR}          & 23.4       & 26.6         & 23.9         & 18.4       & 15.4         & \underline{28.6}         & \underline{30.4}         & \underline{44.9}       & 44.7         & 25.7         & \underline{72.3}         & 44.5       & 14.8         & 47.0         & \underline{56.1}         & 34.5      \\
\rowcolor[HTML]{EBF8FF}
\multicolumn{1}{l|}{\quad$\bullet$ SEVA (Ours)} & 
$\underline{27.4}_{\pm 0.4}$ & 
$\underline{31.0}_{\pm 0.3}$ & 
$\underline{28.1}_{\pm 0.5}$ & 
$\underline{19.5}_{\pm 0.1}$ & 
$\underline{17.9}_{\pm 0.3}$ & 
$\textbf{31.9}_{\pm 0.1}$ & 
$\textbf{33.9}_{\pm 0.3}$ & 
$\textbf{48.2}_{\pm 0.2}$ & 
$46.8_{\pm 0.2}$ & 
$\underline{39.1}_{\pm 0.3}$ & 
$\textbf{72.9}_{\pm 0.0}$ & 
$\textbf{47.1}_{\pm 0.1}$ & 
$13.2_{\pm 1.2}$ & 
$\textbf{50.4}_{\pm 0.0}$ & 
$\textbf{57.1}_{\pm 0.1}$ & 
$\textbf{37.6}_{\pm 0.1}$ \\
\toprule[1pt]
ViT(LN)                                                                              & 9.5        & 6.7          & 8.2          & 29.0       & 23.4         & 33.9         & 27.1         & 15.9       & 26.5         & 47.2         & 54.7         & 44.1       & 30.5         & 44.5         & 47.8         & 29.9      \\
\multicolumn{1}{l|}{\quad$\bullet$ MEMO}      & 21.6       & 17.3         & 20.6         & 37.1       & 29.6         & 40.4         & 34.4         & 24.9       & 34.7         & 55.1         & 64.8         & 54.9       & 37.4         & 55.4         & 57.6         & 39.1       \\
\multicolumn{1}{l|}{\quad$\bullet$ DDA}        & 41.3       & \textbf{41.1}         & 40.7         & 24.4       & 27.2         & 30.6         & 26.9         & 18.3       & 27.5         & 34.6         & 50.1         & 32.4       & 42.3         & 52.2         & 52.6         & 36.1      \\
\multicolumn{1}{l|}{\quad$\bullet$ Tent}       & \underline{42.2}       & 1.0          & \underline{43.3 }        & 52.4       & 48.2         & 55.5         & 50.5         & 16.5       & 16.9         & 66.4         & \underline{74.9}         & 64.7       & 51.6         & 67.0         & 64.3         & 47.7      \\
\multicolumn{1}{l|}{\quad$\bullet$ EATA}       & 29.7       & 25.1         & 34.6         & 44.7       & 39.2         & 48.3         & 42.4         & 37.5       & 45.9         & 60.0         & 65.9         & 61.2       & 46.4         & 58.2         & 59.6         & 46.6     \\
\multicolumn{1}{l|}{\quad$\bullet$ RDumb}       & 36.7       & 32.4         & 37.5         & 49.7       & 46.3         & 55.6         & 49.7         & 53.3       & 45.1         & 63.0         & 71.7         & 61.0       & 53.7         & 66.7         & 61.6         & 52.3      \\
\multicolumn{1}{l|}{\quad$\bullet$ SAR}     & 40.8       & 36.4         & 41.5         & \underline{53.7}       & \underline{50.7}         & \underline{57.5}         & \underline{52.8}         & \underline{59.1}       & \underline{50.7}         & \underline{68.1}         & 74.6         & \underline{65.7}       & \underline{57.9}         & \underline{68.9}         & \underline{65.9}         & \underline{56.3}      \\
\rowcolor[HTML]{EBF8FF}
\multicolumn{1}{l|}{\quad$\bullet$ SEVA (Ours)} & 
$\textbf{44.5}_{\pm 0.3}$ & 
$\underline{40.0}_{\pm 0.9}$ & 
$\textbf{45.3}_{\pm 0.3}$ & 
$\textbf{55.9}_{\pm 0.0}$ & 
$\textbf{53.8}_{\pm 0.2}$ & 
$\textbf{60.4}_{\pm 0.1}$ & 
$\textbf{56.5}_{\pm 0.1}$ & 
$\textbf{62.7}_{\pm 0.1}$ & 
$\textbf{63.1}_{\pm 0.1}$ & 
$\textbf{71.6}_{\pm 0.1}$ & 
$\textbf{77.3}_{\pm 0.0}$ & 
$\textbf{67.1}_{\pm 0.0}$ & 
$\textbf{62.8}_{\pm 0.0}$ & 
$\textbf{71.5}_{\pm 0.1}$ & 
$\textbf{68.5}_{\pm 0.0}$ & 
$\textbf{60.1}_{\pm 0.2}$ \\
\bottomrule[1pt]
\end{tabular}
}
}
\end{table*}

\subsection{Evaluation on Imbalanced Label Shifts}
\label{sec:4.2}
To show the effectiveness of SEVA, we conduct evaluations in scenarios where the class imbalance ratio is infinity and compare it with prior arts such as diffusion-based DDA \cite{DDA}, self-distillation method EcoTTA \cite{ecotta}, and selection methods EATA \cite{eata} and SAR \cite{sar}. To maintain fairness, we refrain from comparing with EcoTTA in ViT evaluation, which is designed for convolutional networks. As detailed in Tab. \ref{tab:label shift}, selection methods EATA and SAR effectively exclude harmful samples to exhibit promising performances in certain corruption cases. However, as discussed in Sec. \ref{sec:selection}, when faced with cases containing a high proportion of confusing samples, these methods struggle to correctly identify the adaptation direction, leading to performance degradation even lower than the no-adapt model, such as the case of ResNet in Defocus and Frost. In contrast, our method benefits from Augmented Entropy-based selection and emerges as the sole method to achieve improvements in these challenging cases, reaffirming the reliability of our selection mechanism. In addition, due to the enhanced adaptation efficiency, SEVA exhibits significantly better performance compared to other methods. Specifically, SEVA showcases its superiority across 11 corruption types for ResNet and all 15 types for ViT, achieving an average performance gain of $\textbf{+4.9, +5.3}$ compared to SAR. Our experimental results underscore the potential of leveraging reliable samples and validate the effectiveness of our method in integrating multi-step augmentation through our novel loss function.

\vspace{-0.1cm}
\subsection{Evaluation on Mixed Testing Domain}
We evaluate the performance of different methods on a mixture of 15 corruption types (15$\times$50,000 images) at severity levels 5, 4, \& 3. As shown in Tab. \ref{table:mixed}, Tent \cite{tent} often encounters a performance collapse and fails to adapt, resulting in lower accuracy compared to the no-adapt model. On the contrary,  both the selection method SAR and the periodic reset method RDumb \cite{rdumb} perform competitively in this long-term test scenario, due to their mechanisms designed for continually changing corruptions. For instance, RDumb incorporates a regular reset mechanism to prevent model collapse and achieve promising results. Nevertheless, the limitation of the adaptation efficiency still hinders their performance. Compared to them, SEVA achieves a consistent improvement on all corruption levels, with an average gain of $\textbf{+2.0, +1.5, +1.0}$ respectively at levels 5, 4, \& 3, demonstrating increasingly stable improvements against stronger corruptions. These results validate the effectiveness of SEVA for handling continually changing corruptions.

\begin{table}[b]
\setlength{\abovecaptionskip}{-0cm} 
\setlength{\belowcaptionskip}{-0.3cm}
\begin{center}
\caption{Efficiency comparison of various methods. We assess TTA approaches for processing 50,000 images in Gaussian type at severity level 5. The practical runtime is evaluated using a single Nvidia RTX 4090 GPU on ResNet50-GN.}
\label{table:running time}
\resizebox{\linewidth}{!}{
\fontsize{17}{22}\selectfont
\begin{tabular}{c|c|ccc|c}
\toprule[1pt]
Method     & Source data & Forward          & Backward         & Other computation  & Time \\
\toprule[1pt]
No-adapt   & \ding{55}            & 50,000           & N/A              & N/A                & 59s       \\
MEMO       & \ding{55}            & 50,000$\times$65 & 50,000$\times$64 & AugMix             & 53,259s    \\
DDA        & \ding{52}            & -                & -                & Diffusion model    & 13,277s    \\
Tent       & \ding{55}            & 50,000           & 50,000           & N/A                & 76s       \\
EATA       & \ding{52}            & 50,000           & 19,608            & regularizer        & 84s       \\
EcoTTA     & \ding{52}            & -                & -                & Meta network       & 11,097s    \\
SAR        & \ding{55}           & 66,418           & 30,488            & More updates & 115s      \\
SAR+VA$\times 5$ &  \ding{55} & 325,070 & 149,244 & Data augmentation & 563s \\
SAR+VA$\times 7$ &  \ding{55} & 454,867 & 214,203 & Data augmentation & 804s \\
\rowcolor[HTML]{EBF8FF}

SEVA(Ours) & \ding{55}            & 50,000           & 18,970            & Eq. \ref{eq:loss}              & 103s    \\
\bottomrule[1pt]
\end{tabular}

}
\end{center}
\end{table}

\subsection{Evaluation on Limited Batch Size}
As depicted in Tab. \ref{tab:bs1}, DDA achieves stable performance on three corruption types of the Noise category, while most other methods suffer performance declines compared to the first two settings. Due to its input adaptation approach, DDA is not affected by limited batch size and maintains the same results. However, the utilization of the diffusion model incurs significant computational overhead, with GPU time tens of times higher than SEVA (shown in Tab. \ref{table:running time}), and only brings limited performance improvements. In such extreme conditions of data streaming input, our method continues to show consistent improvements, especially in ViT evaluation, achieving the best accuracy across all types and an average gain of $\textbf{+3.8}$ compared to the prior SOTA. The
results demonstrate the robustness of our SEVA in the face of data limitations.

\subsection{Running Time Comparison}
In this subsection, we compare the time costs of various TTA methods. As shown in Tab. \ref{table:running time}, DDA and EcoTTA require significantly more time since they need to use the source data to train additional networks to assist in adaptation. Due to the different trained network with other methods, we use $-$ to denote ignoring the number of their forward and backward steps to maintain fairness. MEMO requires numerous additional augmentations to obtain robust predictions, leading to unacceptable time costs and demonstrating that explicit augmentation cannot meet the real-time requirement of the TTA application. In contrast, SEVA does not increase any additional augmentations or training iterations, using less than $2\%$ of the time cost of MEMO. Furthermore, our method maintains a slight advantage of performance compared to SAR with 7 rounds of vicinal augmentation (SAR+VA $\times 7$), and only requires one-eighth of the running time, highlighting the efficiency of our integrated augmentation approach.

\begin{table*}[t]
\setlength{\abovecaptionskip}{-0.0cm} 
\setlength{\belowcaptionskip}{-0.2cm}
\caption{Effects of components in SEVA. We report the Accuracy (\%) on ImageNet-C (level 5) under Imbalanced label shifts.}
\centering
\label{tab:component}
\resizebox{\textwidth}{!}{
{
\fontsize{15}{17}\selectfont
\begin{tabular}{c|ccc|cccc|cccc|cccc|c}
\toprule[1pt]
\multirow{2}{*}{\centering\textbf{Model+Method}}      & \multicolumn{3}{c|}{Noise}              & \multicolumn{4}{c|}{Blur}                          & \multicolumn{4}{c|}{Weather}                       & \multicolumn{4}{c|}{Digital}                       &     \multirow{2}{*}{\centering\textbf{Average}}       \\
    & Gauss.       & Shot       & Impul.     & Defoc.     & Glass      & Motion     & Zoom       & Snow       & Frost      & Fog        & Brit.      & Contr.     & Elastic    & Pixel      & JPEG       &         \\
\toprule[1pt]
\multicolumn{1}{c|}{ResNet50(GN)+Entropy}                                                     & 3.2         & 4.1       & 4.0       & 17.1       & 8.5       & 27.0       & 24.4       & 17.9       & 25.5       & 2.6       & 72.1       & 45.8       & 8.2       & 52.2       & 56.2       & 24.6       \\
\multicolumn{1}{l|}{\quad\ding{58}\, Selection}       & \underline{35.5}         & \underline{37.7}       & \underline{35.6}       & \underline{19.9}       & \underline{21.1}       & \underline{34.0}       & 33.8       & \underline{42.3}       & \underline{45.6}       & \underline{51.2}       & 71.7       & 46.9       & \underline{8.5}       & 52.2       & 56.5       & \underline{39.5}       \\
\multicolumn{1}{l|}{\quad\ding{58}\, $\mathcal{L}_{AE}$}          & 3.9         & 4.7       & 5.2       & 19.0       & 20.9       & 33.2       & \underline{36.6}       & 15.6       & 30.5       & 2.3       & \textbf{72.6}       & \underline{48.1}       & 6.6       & \underline{54.5}       & \underline{57.1}       & 28.2       \\
\rowcolor[HTML]{EBF8FF}
\multicolumn{1}{l|}{\quad\ding{58}\, Selection$+\mathcal{L}_{AE}$}   & \textbf{38.7}  & \textbf{40.8} & \textbf{39.5}  & \textbf{21.7} & \textbf{23.2} & \textbf{37.1} & \textbf{38.0} & \textbf{48.7}  & \textbf{48.1} & \textbf{53.5} & \underline{72.5} & \textbf{48.7} & \textbf{8.7}  & \textbf{54.9} & \textbf{57.2} & \textbf{42.1} \\
\toprule[1pt]
ViT(LN)+Entropy                                                                               & 21.2          & 1.9        & 38.6        & 54.8       & 52.7       & 58.5       & 54.2       & 10.1       & 14.7       & 69.6       & 76.3       & 66.3       & 59.2       & 69.7       & 66.8       & 47.6       \\
\multicolumn{1}{l|}{\quad\ding{58}\, Selection}       & 47.5         & \underline{45.9}       & 48.0       & 51.7       & 48.7       & 57.3       & 49.7       & 56.8       & 55.1       & 60.4       & 68.4       & 59.4       & 53.0       & 60.7       & 61.2       & 54.9       \\
\multicolumn{1}{l|}{\quad\ding{58}\, $\mathcal{L}_{AE}$}          & \underline{49.9}        & 28.1       & \underline{50.9}       & \underline{56.7}       & \underline{56.9}      & \underline{61.9}       & \underline{57.9}       & \underline{65.1}       & \underline{64.6}       & \underline{72.1}       & \textbf{77.4}       & \underline{67.6}       & \underline{65.4}       & \underline{71.9}       & \underline{68.8}       & \underline{61.0}       \\
\rowcolor[HTML]{EBF8FF}
\multicolumn{1}{l|}{\quad\ding{58}\, Selection$+\mathcal{L}_{AE}$} & \textbf{52.2}   & \textbf{51.9} & \textbf{53.1} & \textbf{57.5} & \textbf{57.9} & \textbf{62.3} & \textbf{59.3} & \textbf{66.3} & \textbf{65.0} & \textbf{72.4} & \textbf{77.4} & \textbf{67.7} & \textbf{66.6} & \textbf{72.3} & \textbf{69.2} & \textbf{63.4} \\
\bottomrule[1pt]
\end{tabular}
}
}
\end{table*}

\section{Ablation Study and Analysis}
In this section, without loss of generality, we conduct ablation studies on imbalanced label shifts in Sec. \ref{sec:4.2} for the sake of brevity. Focusing on two pivotal components of SEVA, the novel loss $\mathcal{L}_{AE}$ that integrates augmentations and the reliable sample selection mechanism, we perform various experiments to analyze their impacts, aiming to delve into the influence of hyperparameters and implementation details, gaining insights into key factors contributing to the effectiveness of SEVA.

\subsection{Effectiveness of Components in SEVA}
To validate the effects of pivotal components in our method, we compare our method with pure entropy minimization and independently add parts of our method to study their impact on performance. As shown in Tab. \ref{tab:component}, replacing vanilla entropy with Augmented Entropy can consistently improve adaptation performance, with an increase $47.6\% \to 61.0\%$ on ViT-LN and $24.6\% \to 28.2\%$ on ResNet50-GN. This simple replacement leads to improvements in almost all corruption types, especially achieving significant improvements for some corruption types on ViT-LN, such as Gaussian (+28.7), Snow (+55.0), and Frost (+49.9). Our reliable selection mechanism also shows the expected effect of reducing the risk of collapse, obtaining improvements $47.6\% \to 54.9\%$ on ViT-LN and $24.6\% \to 39.5\%$ on ResNet50-GN. In cases where entropy fails, such as the three corruption types of Noise, our selection mechanism achieves improvements of $+32.3, +33.6, \text{and} +31.6$, respectively. Furthermore, in all scenarios, the complete SEVA (Selection + $\mathcal{L}_{AE}$) achieves the best performance, indicating that the two key components are compatible with each other, further enhancing adaptation.

\subsection{Comparison on Selection Mechanism}
\begin{figure}[b]
\setlength{\abovecaptionskip}{-0.2cm} 
\setlength{\belowcaptionskip}{-0.2cm}
\begin{center}
\includegraphics[width=1.0\linewidth]{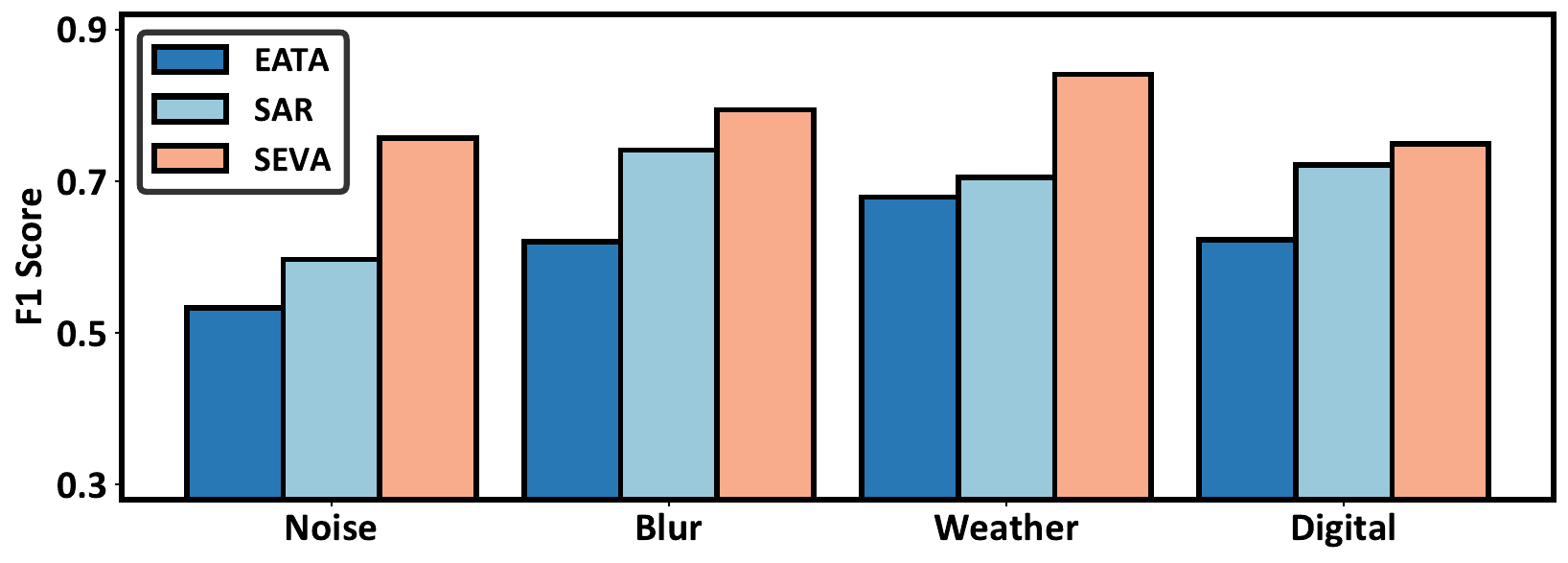}
\end{center}
\caption{Comparison of various sample selection methods on the ViT-LN model. A large F1 score indicates that the method can identify reliable samples effectively.}
\label{fig:selection}
\end{figure}

We have analyzed the advantages of Augmented Entropy over vanilla entropy for selection in Sec. \ref{sec:selection}. In this section, we provide a direct comparison to validate the effectiveness of our selection mechanism. First, we propose a simple criterion of reliable samples: if the prediction for a testing sample matches its ground-truth class, then the optimization of its entropy-based training will move towards the correct one-hot label direction. In this case, we can intuitively consider it as a reliable sample. We use this criterion to calculate the F1 score of different selection strategies, as shown in Fig. \ref{fig:selection}. Results for all four categories validate the superiority of our selection mechanism over other methods. For the weather category, our F1 score increases by nearly $+0.15$ compared to SAR, bringing the significant accuracy improvement of $+8.9\%$. This outcome further demonstrates the importance of our selection mechanism.

\vspace{-0.1cm}
\subsection{Comparison with Augmentation}
As shown in Fig. \ref{fig:compare}, we compare various augmentation methods and different numbers of augmentation rounds. The results reveal that training with augmentation can consistently improve model performance and the effectiveness of different augmentations varies significantly. Vicinal augmentation and AugMix achieve better performance, while Manifold Mix and CutMix show limited improvement due to the lack of ground-truth labels to mix. Since vicinal augmentation performs the best and provides a concise formula that facilitates our
theoretical framework, we leverage it as the augmentation for SEVA integration. Furthermore, as shown in Fig. \ref{fig:compare}(c), SEVA achieves supreme performances surpassing those of direct augmentation for $7$ times without the need for explicit augmentation, demonstrating the remarkable improvement of SEVA in enhancing adaptation efficiency.

\begin{figure}[h]
\setlength{\belowcaptionskip}{-0.8cm} 
\setlength{\abovecaptionskip}{-3pt} 
\begin{center}
\includegraphics[width=1.0\linewidth]{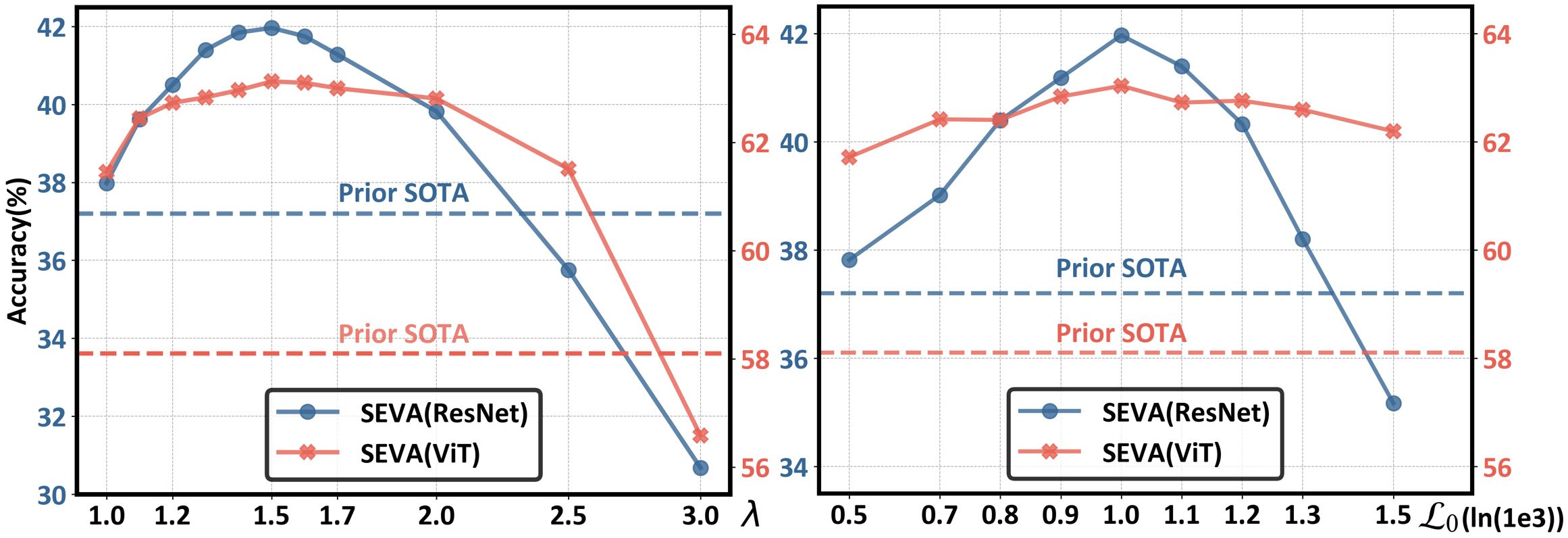}
\end{center}
\caption{The left picture shows the performance under different vicinal range $\lambda$ and the right shows the performance under different boundary coefficient $\mathcal{L}_0$.}
\label{fig:hyper}
\end{figure}

\vspace{-0.2cm}
\subsection{Hyperparameter Robustness}
For the two components of our method, there are two important hyper-parameters: the coefficient $\lambda$ for the range of vicinal area and the coefficient $\mathcal{L}_0$ for the boundary. We conduct ablation experiments on these two key coefficients independently:

$\lambda$ determines the vicinal range for augmentation by influencing the scale of the variance. Specifically, as $\lambda$ increases, the variance of the Gaussian distribution increases, expanding the vicinal range. In Fig. \ref{fig:hyper}, we evaluate our method under various range settings. It is evident that performance improves as the vicinal area expands within a specific range and peaks at $\lambda=1.5$. However, if the vicinal area exceeds a normal range (e.g., 3.0), the noise will destroy sample semantics, causing harmful optimization and performance degradation. Therefore, we set the coefficient $\lambda$ to 1.5 by default. Within the range of $\lambda \in [1.0,2.0]$, the performance consistently outperforms prior SOTA, demonstrating the robustness of our SEVA.

$\mathcal{L}_0$ serves as a boundary to filter out harmful samples. Specifically, as $\mathcal{L}_0$ decreases, the number of excluded samples increases, ensuring a more reliable selection but also reducing the frequency of model updates. In converse, as $\mathcal{L}_0$ increases, more samples are involved in training, increasing the update frequency but reducing the accuracy of the adaptation direction. As shown in Fig. \ref{fig:hyper}, when $\mathcal{L}_0$ is too large or too small, the balance between update frequency and adaptation accuracy is broken, resulting in a performance decrease. Within the range $\mathcal{L}_0 \in [0.7,1.2]$, the performance consistently exceeds the latest SOTA by $+2.0\%$, indicating the robustness of SEVA to the boundary. To ensure efficiency and stability, we set $\mathcal{L}_0$ to $1.0 \times \ln(1000)$ in all experiments.
\vspace{-0.3cm}
\section{Conclusion and Future Work}
In this work, we focus on unleashing the power of augmentation strategies in leveraging the potential of reliable samples for Test-Time adaptation. We develop a theoretical analysis of how vicinal augmentation affects the entropy-based training process and propose an efficient loss to integrate multi-step augmentation training into a single step, significantly boosting adaptation efficiency under the stringent time requirements of TTA. Meanwhile, our novel loss is conducive for sample selection to exclude harmful ones and can further improve the accuracy of adaptation direction. Our method showcases consistent effectiveness across a series of challenging adaptation scenarios. Similar to Tent and other TTA methods compatible with diverse multimedia scenarios, our approach, as an enhancement of the entropy-based training framework, also has the potential to be applied in a broader range of multimedia settings. We intend to broaden our evaluation in future work to delve deeper into these possibilities of our method.

\bibliographystyle{ACM-Reference-Format}
\bibliography{sample-base}

\end{document}